\documentclass{article}
\usepackage{spconf,amsmath,graphicx,hyperref}
\usepackage{cite}
\usepackage{booktabs}
\usepackage{amsmath}
\usepackage{tikz}
\usepackage{amsfonts}
\usepackage{hyperref}
\usetikzlibrary{arrows.meta,positioning}

\title{Emotion-Aligned Generation in Diffusion Text to Speech Models via Preference-Guided Optimization}
\name{Jiacheng Shi$^{\star}$ \qquad Hongfei Du$^{\star}$ \qquad Yangfan He$^{\dagger}$ \qquad Y. Alicia Hong$^{\ddagger
}$ \qquad Ye Gao$^{\star}$}
\address{$^{\star}$ College of William \& Mary, $^{\dagger}$ University of Minnesota - Twin Cities, $^{\ddagger}$ George Mason University\\
\small \texttt{\{jshi12, hdu02, ygao18\}@wm.edu, he000577@umn.edu, yhong22@gmu.edu}}

\begin{document}
%
\maketitle
\begin{abstract}
Emotional text-to-speech seeks to convey affect while preserving intelligibility and prosody, yet existing methods rely on coarse labels or proxy classifiers and receive only utterance-level feedback. We introduce Emotion-Aware Stepwise Preference Optimization (EASPO), a post-training framework that aligns diffusion TTS with fine-grained emotional preferences at intermediate denoising steps. Central to our approach is EASPM, a time-conditioned model that scores noisy intermediate speech states and enables automatic preference pair construction. EASPO optimizes generation to match these stepwise preferences, enabling controllable emotional shaping. Experiments show superior performance over existing methods in both expressiveness and naturalness.
\end{abstract}

\section{Introduction}
\label{sec:intro}
 Emotional text-to-speech (TTS) \cite{chen2022fine,guo2023prompttts,diatlova2023emospeech,guo2023emodiff,du2024cosyvoice,du2024cosyvoice2, yang2025emovoice} aims to generate speech that remains intelligible while conveying affect and prosodic nuance. Effective emotional control supports conversational agents, accessibility, and content creation. Yet fine-grained control without sacrificing naturalness remains challenging, as emotional cues unfold over time and interact with linguistic content and speaker variation.

Early emotional TTS systems fused local style tokens with phoneme content via cross-attention for fine-grained control~\cite{chen2022fine}, guided synthesis with semantic prompts~\cite{guo2023prompttts}, and used variational decoders to model nuanced emotion–prosody relations~\cite{diatlova2023emospeech}. Diffusion-based methods improved fidelity and control through emotion embeddings and noise-conditioned prosody~\cite{guo2023emodiff}. Recent work adopts supervised speech tokens for semantic grounding and chunk-aware flow-based decoding~\cite{du2024cosyvoice,du2024cosyvoice2}, and introduces parallel phoneme–audio branches in LLM-based generation to support fine-grained, freestyle emotional control~\cite{yang2025emovoice}. Across these variants, two important gaps remain: supervision still targets proxy labels over preference-driven prompts, and feedback remains temporally sparse, limiting constraints on prosody–emotion dynamics.
Direct Preference Optimization (DPO)\cite{rafailov2023direct} aligns generative models to human choices via paired comparisons against a frozen reference policy. In diffusion settings\cite{wallace2024diffusion}, a common practice is to assign preference at the final step and propagate it to intermediate latents, then optimize a DPO-style log-likelihood ratio at each step. This removes the need for an explicit reward model and is effective for global preferences, but offers limited guidance for temporally evolving signals.
DPO has also been used to align large language model–based TTS. EmoDPO~\cite{gao2025emo}, for instance, pairs utterances with identical text and marks the target emotion as preferred, achieving alignment without explicit reward modeling. However, attaching preference only at the endpoint yields sparse supervision for gradually varying cues, the assumption that all intermediate states on a preferred path are themselves preferred is often invalid, and curating domain-appropriate preference pairs is expensive. This leads to our central research question: \textit{How can we provide dense, fine-grained supervision for emotionally expressive TTS generation, without relying on endpoint preferences or categorical labels?}

To better align diffusion-based TTS systems with fine-grained emotional preferences, we propose \textit{Emotion-Aware Stepwise Preference Optimization} (EASPO). At each denoising step starting from a latent representation $x_t$, the model samples a small candidate set of $x_{t-1}$ mel-spectrograms. An \textit{Emotion-aware Stepwise Preference Model} (EASPM) scores their emotional expressiveness and selects a win--lose pair that differs subtly in prosody while preserving linguistic content. One candidate is then randomly chosen to continue generation. Since these candidates originate from the same latent and differ by only a single denoising step, their variations are localized and emotion-focused. EASPM captures these nuanced differences and steers the model toward producing more emotionally consistent speech.
Our contributions are: 1) We propose EASPO, a stepwise alignment framework that reformulates preference optimization as a local, time-conditioned task, replacing the flawed assumption that all intermediate states on a preferred trajectory are equally valid. By aligning win/lose candidates at each denoising step from a shared latent, it enables stepwise-controllable emotion shaping throughout the generation process. 2) We introduce EASPM, a time-aware reward model that directly scores emotional expressiveness and prosody on noisy intermediate states, enabling dense, temporally grounded preference learning and on-the-fly scoring.

\begin{figure*}[htbp]
    \centering
    \vspace{-14mm}
    \includegraphics[width=0.8\textwidth]{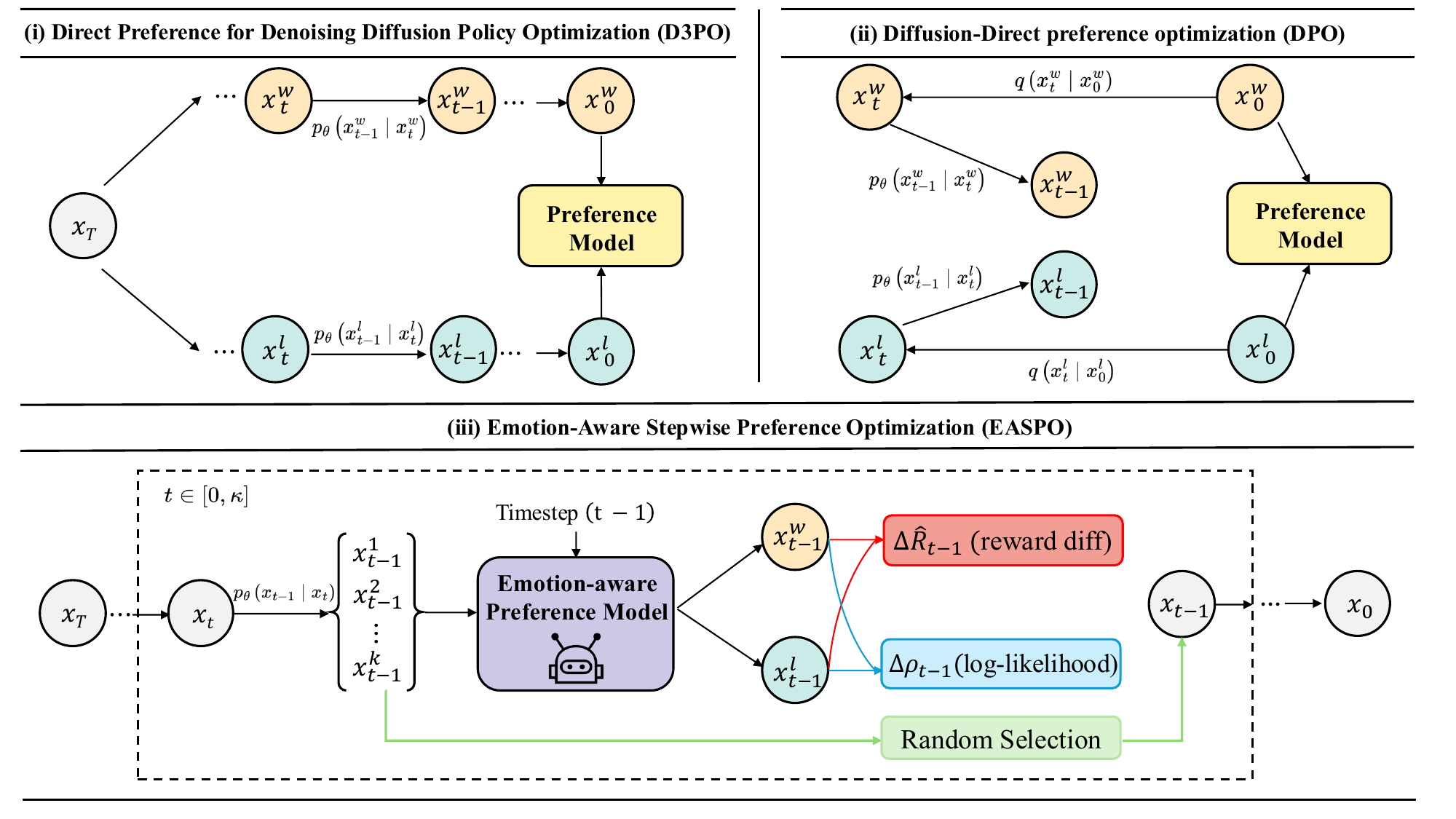}
    \vspace{-4mm}
    \caption{Unlike prior DPO-based methods, EASPO avoids direct propagation of preferences across diffusion steps. At each step in EASPO, a set of candidate samples is produced, from which a suitable win–lose pair is chosen to update the diffusion model. Afterward, one sample is randomly picked to serve as the starting point for the next iteration.}
    \label{fig:main}
    \vspace{-2mm}
\end{figure*}
    \section{Method}
We present a reinforcement learning framework for emotional TTS that fine-tunes diffusion models via stepwise preference supervision (Fig.\ref{fig:main}). Starting from a pretrained Grad-TTS\cite{popov2021grad}, our method introduces dense emotion-aligned rewards at each denoising step. At every latent state, multiple mel-spectrograms are sampled and ranked by a frozen emotion preference model(Sec.~\ref{sec:EASPM}). A preference pair is selected, and the model is optimized to favor the emotionally preferred sample by minimizing the gap between its advantage and the log-likelihood ratio(Sec.~\ref{sec:easpo-objective}). Our EASPO objective extends DPO with stepwise preference signals, enabling fine-grained emotional control during generation.

\subsection{Emotion-Aware Stepwise Preference Model}
\label{sec:EASPM}

\textbf{Overall.}
At reverse step $t-1$, given the current latent $x_t$, the generator draws $k$ candidates
$\{x_{t-1}^1,\ldots,x_{t-1}^k\}\sim p_\theta(x_{t-1}\mid x_t)$. EASPM assigns each candidate a \emph{timestep-aware} emotion–prompt consistency score and induces a ranking; the highest and lowest scored items are taken as the win/lose pair.

\noindent\textbf{Scoring.} EASPM is built on CLEP~\cite{shi2025clep}, a CLAP-based~\cite{elizalde2023clap} contrastive audio–language encoder fine-tuned on large-scale emotional speech data.
Let $f_{\text{CLEP-A}}(\cdot,t)$ and $f_{\text{CLEP-T}}(\cdot)$ denote the audio and text branches.
Using $\ell_2$-normalized embeddings, the score for candidate $x_{t-1}^i$ is
\begin{equation}
s_i \;=\;
\Big\langle
\frac{f_{\text{CLEP-A}}(x_{t-1}^{i},t)}{\big\|f_{\text{CLEP-A}}(x_{t-1}^{i},t)\big\|_2},\,
\frac{f_{\text{CLEP-T}}(c)}{\big\|f_{\text{CLEP-T}}(c)\big\|_2}
\Big\rangle .
\label{eq:EASPM-score}
\end{equation}
For a pair $(x_{t-1}^{w},x_{t-1}^{l})$ with scores $s_w,s_l$, define $\Delta_t=s_w-s_l$.
The probability that the win item is preferred is modeled by a pairwise logistic with temperature $\tau>0$:
\begin{equation}
\hat p_w \;=\; \sigma(\tau\,\Delta_t)
\;=\; \frac{1}{1+\exp(-\tau\,\Delta_t)} .
\label{eq:EASPM-prob}
\end{equation}
Candidate selection is performed by
$x_{t-1}^{\text{win}}=\arg\max_i s_i$ and
$x_{t-1}^{\text{lose}}=\arg\min_i s_i$, and the stepwise preference loss is:
\begin{equation}
\mathcal{L}_{\text{pref}}
\;=\; -\log \hat p_w
\;=\; \log\!\big(1+\exp(-\tau\,\Delta_t)\big).
\label{eq:EASPM-loss}
\end{equation}

\noindent\textbf{Training.} 
EASPM is adapted from CLEP to handle noisy intermediate representations. Given a win–lose pair \((x_0^w, x_0^l)\), we sample a timestep \(t\) and apply the same forward diffusion to produce \((x_t^w, x_t^l)\). This tuple \((x_t^w, x_t^l, t, c)\) is used to optimize \(\mathcal{L}_{\text{pref}}\), encouraging the model to recover the correct preference at step \(t\). A time-aware normalization layer is added to CLEP's audio branch for timestep conditioning. To reduce mismatch with CLEP’s pretraining domain, we optionally estimate a pseudo-clean \(\hat{x}_0\) from \(x_t\) via deterministic inversion followed by~\cite{song2020denoising} . After training, EASPM is frozen and used solely as a stepwise scorer for the RL objective.

\noindent\textbf{Random selection of the next state.}
After EASPM ranks the candidate set $\{x_{t-1}^1,\ldots,x_{t-1}^k\}$ and selects a win–lose pair, we \emph{do not} continue with the top sample as shown in Fig.~\ref{fig:main}. To avoid biased rollouts and degenerate paths, we uniformly sample the next state $\tilde{x}_{t-1}$ from the pool and proceed with $x_{t-2} \sim p_\theta(\cdot \mid \tilde{x}_{t-1}, c)$. This ensures all preference pairs originate from the same latent $x_t$ while decoupling supervision from sampling. Candidate pooling is applied only when $t \le \kappa$; standard transitions are used for $t > \kappa$.

\subsection{Objective Function of EASPO}
\label{sec:easpo-objective}

We formulate denoising as a $T$-step MDP with state $s_t=(c,x_t)$, action $a_t=x_{t-1}$, and policy
$\pi_\theta(a_t\mid s_t)=p_\theta(x_{t-1}\mid x_t,c)$.
At each step $t$, we sample $k$ candidates $\{x_{t-1}^i\} \sim p_\theta(\cdot \mid x_t, c)$
and rank them via EASPM (Sec.~\ref{sec:EASPM}). Let $x_{t-1}^{w}$ and $x_{t-1}^{l}$ denote the top and bottom-ranked samples from the \emph{same} latent $x_t$.

\noindent\textbf{Dense stepwise reward.}
EASPM supplies a dense emotional reward at step $t$:
\begin{equation}
\widehat{R}_t^{j} \;=\; s\!\left(x_{t-1}^{j},c,t\right), 
\qquad
\Delta\widehat{R}_t \;=\; \widehat{R}_t^{w}-\widehat{R}_t^{l},
\label{eq:easpo-reward}
\end{equation}

\noindent\textbf{Log-likelihood ratio against a reference policy.}
Let the frozen reference be $\pi_{\mathrm{ref}}(a_t\mid s_t)=p_{\theta_{\mathrm{ref}}}(x_{t-1}\mid x_t,c)$. 
For $j\in\{w,l\}$ we define
\begin{equation}
\rho_t^{j}(\theta)
= \log \pi_\theta\!\left(x_{t-1}^{j}\mid s_t\right)
  - \log \pi_{\mathrm{ref}}\!\left(x_{t-1}^{j}\mid s_t\right),
\label{eq:easpo-logratio}
\end{equation}
and use the win–lose difference $\Delta\rho_t=\rho_t^{w}(\theta)-\rho_t^{l}(\theta)$ to measure the policy’s preference change relative to the reference.

\noindent\textbf{Stepwise alignment objective.}
Inspired by~\cite{deng2024prdp}, we align the log-ratio difference with the dense reward difference via a mean-squared error with a time weight
\(
\beta_t=\lambda^{\,T-t-1}/\eta
\) (\(\lambda\!\in\!(0,1],\,\eta\!>\!0\)):
\begin{equation}
\mathcal{L}_{t}(\theta)
=
\big(\,\beta_t\,\Delta \rho_t \;-\; \Delta \widehat{R}_t\,\big)^{2}.
\label{eq:easpo-step}
\end{equation}

\noindent\textbf{Final EASPO loss.}
To improve sample efficiency, we optimize at a randomly shuffled step $\tau$ and skip the first $\kappa$ high-noise steps.
Averaging over prompts $c$, initial noises $x_T$, and win/lose pairs from the policy gives
\begin{equation}
\begin{aligned}
\mathcal{L}(\theta)
&= \mathbb{E}_{c \sim p(c),x_T \sim \mathcal{N}(0,I),\tau \sim \mathcal{U}[1,\,T-\kappa],x_{\tau-1}^{w},\,x_{\tau-1}^{l} \sim p_\theta(\cdot \mid x_\tau, c)}\,
\\[-2pt]
&\!\!\!\!\!\!\!\!\!\!\!\! \bigg[
\Big(
\beta_{\tau}\,\big[ \rho_{\tau}^{w}(\theta) - \rho_{\tau}^{l}(\theta) \big]
\;-\;
\big[ s(x_{\tau-1}^{w}, c, \tau) - s(x_{\tau-1}^{l}, c, \tau) \big]
\Big)^{2}
\bigg],
\end{aligned}
\label{eq:easpo-final-expanded}
\end{equation}


\noindent\textbf{Discussion.}
Eq.~\eqref{eq:easpo-final-expanded} integrates stepwise preference optimization with reward-difference learning:
EASPM provides a \emph{dense emotional reward} (Eq.~\eqref{eq:easpo-reward}), 
and the diffusion policy is updated so that its \emph{log-likelihood ratio difference} between win/lose transitions (Eq.~\eqref{eq:easpo-logratio}) matches that reward difference (Eq.~\eqref{eq:easpo-step}).
Empirical results from~\cite{deng2024prdp} show that using reward differences yields more stable reward optimization in diffusion models than standard policy gradient methods.

\section{Experiments}
\subsection{Datasets and Experimental Setup}
\noindent
We fine-tune EASPM on the English MSP-Podcast corpus~\cite{lotfian2017building} ($\sim$55k utterances, $>$1{,}200 speakers), using textual prompts from emotion labels and acoustic descriptors (e.g., pitch, loudness, jitter, shimmer). Preference pairs are created by labeling target emotions (e.g., happy) as preferred over distractors (e.g., neutral) with the same text. For reinforcement learning, we use the English split of ESD (5 emotions × 10 speakers, 350 utterances/emotion), with an 8:1:1 train/val/test split per speaker–emotion. We evaluate against seven emotion-controllable TTS baselines: FG‑TTS~\cite{chen2022fine}, PromptTTS~\cite{guo2023prompttts}, Emospeech~\cite{diatlova2023emospeech}, EmoDiff~\cite{guo2023emodiff}, CosyVoice~\cite{du2024cosyvoice}, CosyVoice2~\cite{du2024cosyvoice2}, and EmoVoice~\cite{yang2025emovoice}, using authors’ code and checkpoints with default inference settings.
\begin{table}[t]
\vspace{-4mm}
\centering
\caption{Objective comparison of our approach with other emotion-controllable TTS models in terms of emotion similarity, prosody similarity, WER, and UTMOS on ESD dataset.}
\resizebox{\columnwidth}{!}{
\begin{tabular}{lcccc}
\toprule[1pt]
TTS Model            & Emo\_SIM$\uparrow$ & Prosody\_SIM$\uparrow$ & 
 WER$\downarrow$ &
UTMOS$\uparrow$ \\ 
\midrule
FG-TT~\cite{chen2022fine}      & 93.91  & 3.28  & 9.38  & 3.81 \\
PromptTTS~\cite{guo2023prompttts}      & 95.70  & 3.41  & \textbf{3.25}  & 4.33 \\
Emospeech~\cite{diatlova2023emospeech}    & 96.35 & 3.39  & 7.13  & 4.24 \\
EmoDiff~\cite{guo2023emodiff}    & 96.62 & 3.55  & 5.62  & 4.35 \\
CosyVoice~\cite{du2024cosyvoice}     & 97.07  & 3.64  & 4.32  & 4.41 \\
CosyVoice2~\cite{du2024cosyvoice2}     & 98.47  & \underline{3.78}  & 3.83  & \underline{4.43} \\
EmoVoice~\cite{yang2025emovoice}     & \underline{98.59}   & 3.67   & 4.16  & 4.39 \\ \midrule
\textbf{Ours}     & \textbf{99.15}   & \textbf{3.89 }  & \underline{3.74}  & \textbf{4.47} \\ 
\bottomrule[1pt]
\end{tabular}
}
\label{tab:objective}
\end{table}
\begin{table}[t]
\vspace{-4mm}
\centering
\caption{Subjective evaluation of naturalness, emotional expressiveness , emotion consistency, and emotion recall, evaluated by human raters. }
\label{tab:mos}
\resizebox{\columnwidth}{!}{
\begin{tabular}{lcccc}

\toprule
TTS Model & MOS$\uparrow$ & Emo\_MOS$\uparrow$ & MOS\_EC$\uparrow$ &
Recall$\uparrow$
\\
\midrule
PromptTTS~\cite{guo2023prompttts}  & 2.95 & 2.88 & 2.72 & 74.12\\
EmoDiff~\cite{guo2023emodiff} & 3.28 & 3.36 & 3.40 & 78.59\\
CosyVoice2~\cite{du2024cosyvoice2} & \underline{3.63} & 3.71 & \underline{3.83} & \underline{82.10}\\
EmoVoice~\cite{yang2025emovoice}     & 3.56 & \underline{3.79} & 3.64 & 80.36\\

\midrule
\textbf{Ours}  & \textbf{3.94} & \textbf{4.28} & \textbf{4.04} & \textbf{85.84}\\
\bottomrule
\end{tabular}
}
\vspace{-4mm}
\end{table}

 We initialize EASPM from CLEP. Audio is resampled to 16\,kHz and cropped/padded to 5\,s. The text encoder is frozen, while the audio encoder and projection head are trained using Adam (batch size 64, 80 epochs), with learning rates $1\!\times\!10^{-5}$ and $1\!\times\!10^{-3}$, respectively. Preference supervision is derived from emotion-labeled recordings, marking the target emotion as preferred and another as dis-preferred. To make scoring step-aware, both waveforms in a pair are perturbed by identical diffusion noise at a sampled denoising step. Our base TTS model is Grad-TTS with 80-dim mel-spectrograms. We freeze the encoder and duration predictor and fine-tune only the decoder (score network), initialized from Grad-TTS pretraining settings (Adam, $1\!\times\!10^{-4}$ LR, batch size 16, random 2\,s mel segments). During EASPO, we apply step shuffling and candidate pooling. A denoising step $\tau \sim \mathcal{U}(1, T{-}\kappa)$ is chosen (skipping noisy early steps), $k{=}4$ candidates are sampled from $p_\theta(\cdot \mid x_\tau, c)$, ranked by EASPM, and a win–lose pair is used to minimize the stepwise difference–matching loss between policy log-ratio difference and reward difference. Unless specified, we set $\kappa{=}0.25T$, batch size 32, and decoder learning rate $1\!\times\!10^{-5}$. Waveforms are synthesized using a pretrained HiFi-GAN vocoder~\cite{kong2020hifi}.

\begin{table*}[t]
\centering
\vspace{-8mm}
\begin{minipage}[t]{0.33\linewidth}
\setlength{\tabcolsep}{1.08mm}
\centering
\footnotesize
\caption{{Comparing EASPM with variants: no time condition.}} 
\label{tab:SPM}
{
\begin{tabular}{lcccc}
\toprule[1pt]
Prefer. model     &  E-S & P-S & WER & UTMOS\\
\midrule
    {EASPM}  & \textbf{99.15} & \textbf{3.89} & \textbf{3.74} & \textbf{4.47} \\
    {w/o step con.}   &  98.79 & 3.81 &  3.83 & 4.36 \\
    {CLAP} &95.84 &3.36 &3.96 &4.05\\
    \bottomrule[1pt]
\end{tabular}
}
\end{minipage}
\begin{minipage}[t]{0.33\linewidth}
\setlength{\tabcolsep}{1.68mm}
\centering
\footnotesize
\caption{{Comparing random sampling with other sampling strategies.}} 
\label{tab:reinitialization}  
{
\begin{tabular}{lcccc}  
\toprule[1pt]
Initial.  &  E-S & P-S & WER & UTMOS\\
\midrule
$\boldsymbol{x}^{w}_{t-1}$   &  97.78 & 3.63 & 3.81 & 4.20\\
$\boldsymbol{x}^{l}_{t-1}$   &  98.39 & 3.75 & 3.79 & 4.33 \\
random    & \textbf{99.15} & \textbf{3.89} & \textbf{3.74} & \textbf{4.47} \\
\bottomrule[1pt]
\end{tabular}
}
\end{minipage}
\begin{minipage}[t]{0.33\linewidth}
\setlength{\tabcolsep}{1.08mm}
\centering
\footnotesize
\caption{{Impact of number of sampled images $k$ at each step. We use $k=4$.}}
\label{tab:num_of_sample_per_step_ablation}
{
\begin{tabular}{ccccc}  
\toprule[1pt]
  \#{}samples $k$ &  E-S & P-S & WER & UTMOS\\ 
    \midrule
    2 & 98.31 & 3.76 & 3.78 & 4.23 \\
    4 & \textbf{99.15} & 3.89 & 3.74 & \textbf{4.47}\\
    8 & 98.84 & \textbf{3.93} & \textbf{3.71} & 4.27\\
    \bottomrule[1pt]
\end{tabular}
}
\end{minipage}
\begin{minipage}[t]{0.33\linewidth}
\setlength{\tabcolsep}{0.85mm}
\centering
\footnotesize
\caption{Impact of timestep range.}
\label{tab:timestep_range}
{
\begin{tabular}{ccccc} 
\toprule[1pt]
  Timestep Range  &  E-S & P-S & WER & UTMOS\\  
    \midrule
    \texttt{[0-250]}   &  98.16 & 3.35 & 3.81 & 4.14 \\
    \texttt{[0-500]}   &  98.79 & 3.62 & 3.76 & 4.39 \\
    \texttt{[0-750]}   & \textbf{99.15} & \textbf{3.89} & 3.74 & \textbf{4.47} \\
    \texttt{[0-1000]}   &  97.92 & 3.57 & \textbf{3.69} & 4.26 \\
    \texttt{[250-750]} & 98.85 & 3.81 & 3.71 & 4.42 \\
    \texttt{[500-750]} & 98.34 & 3.69 & 3.75 & 4.37 \\
    \texttt{[250-500]} & 98.68 & 3.77 & 3.73 & 4.28 \\
    \bottomrule[1pt]
\end{tabular}  
}
\end{minipage}
\begin{minipage}[t]{0.33\linewidth}
\setlength{\tabcolsep}{1mm}
\centering
\footnotesize
\caption{Comparison with other diffusion based RL alignment methods.}
\label{tab:align_methods}
{
\begin{tabular}{lcccc}  
\toprule[1pt]
    Method     &  E-S & P-S & WER & UTMOS \\
    \midrule 
    Vanilla-DM    & 96.62 & 3.55  & 5.62  & 4.35 \\
    DDPO      & 98.37 & 3.63 & 4.07 & 4.41 \\
    D3PO      & 97.51 & 3.59 & 4.41 & 4.40   \\
    Diff.-DPO  & 97.85 & 3.67 & 3.82 & 4.37  \\
    EASPO     & \textbf{99.15} & \textbf{3.89} & \textbf{3.74} & \textbf{4.47} \\
    \bottomrule[1pt]

\end{tabular}  
}
\end{minipage}
\begin{minipage}[t]{0.33\linewidth}
\setlength{\tabcolsep}{1.08mm}
\centering
\footnotesize
\caption{{Comparison of win–lose pair selection strategies. Using candidates with highest and lowest emotional scores yields stronger contrast and better alignment than random sampling.}}
\label{tab:win_lose_choice}
{
\begin{tabular}{lcccc}  
\toprule[1pt]
  win-lose sample  &  E-S & P-S & WER & UTMOS\\  
    \midrule  
    best \& worst    SPO     & \textbf{99.15} & \textbf{3.93} & \textbf{3.74} & \textbf{4.47} \\
    random   &  98.82 & 3.89 & 3.95 & 4.36 \\
    \bottomrule[1pt]
\end{tabular}  
}

\end{minipage}
\vspace{-4mm}
\end{table*}
\subsection{Evaluation Metrics}

\noindent\textbf{Objective metrics.} 
\textit{Emo\_SIM} quantifies emotional alignment as the average cosine similarity between emotion2vec-base embeddings of generated and reference utterances. \textit{Prosod\_SIM} is computed via AutoPCP~\cite{barrault2023seamless}, which compares utterance-level prosody (rhythm, stress, intonation). \textit{WER} (word-error-rate) is calculated using Whisper Large-v3 transcripts. 
UTMOS~\cite{saeki2022utmos}  is employed to evaluate speech naturalness and perceptual quality.

\noindent\textbf{Subjective metrics.}
\noindent
We conduct listening tests with 20 raters. Each system is evaluated on 30 clips per rater (six per emotion across five emotions). \textit{MOS} assesses naturalness, and \textit{Emotion MOS} evaluates how well the target emotion is conveyed given the prompt, both on a 1–5 scale (0.5 increments). \textit{MOS\_EC} measures consistency between the generated audio and the instruction (emotion + text). For \textit{Emotion Recall}, raters identify the perceived emotion from five choices; accuracy is averaged over both utterances and emotion classes.

\subsection{Main Results}
To validate the effectiveness of our proposed emotionally preference-aligned method for TTS, we compare it with seven recent emotion-controllable baselines. As shown in Table~\ref{tab:objective}, our method achieves superior performance across emotion similarity, prosody similarity, intelligibility (WER), and perceptual quality (UTMOS), with a notable 2.07\% gain over CosyVoice in Emo\_SIM. A similar trend is observed in Table~\ref{tab:mos}, where our model consistently outperforms baselines in naturalness (MOS), emotional expressiveness (Emo\_MOS), emotion consistency (MOS\_EC), and emotion classification accuracy (Recall). These results demonstrate the effectiveness of our approach in generating emotionally aligned speech with enhanced coherence, while preserving natural prosody and speech quality. Demo page is available~\href{https://jiachengqaq.github.io/emo-tts-demo/}.

\subsection{Ablation Study}

\noindent\textbf{Effectiveness of the Emotion-aware Stepwise Preference Model.}  We ablate timestep conditioning and CLEP initialization to assess their roles in Tab.~\ref{tab:SPM}. Removing either results in consistent performance drops, confirming that both components are essential for accurate step-aware emotional scoring.

\noindent \textbf{Random Selection for Next Iteration Initialization.}  
We compare random selection from the candidate pool with reusing the previous lose sample $x_{t-1}^{l}$ to initialize the next denoising step in Tab.~\ref{tab:reinitialization}. Random selection consistently improves overall performance by avoiding bias toward dispreferred regions and encouraging trajectory diversity.

\noindent\textbf{Impact of Candidate Pool Size.}  
We vary the number of candidates k at each step and observe its effect in Tab.~\ref{tab:num_of_sample_per_step_ablation}. A moderate k balances contrastive supervision and fidelity, improving the learning of emotional and prosodic preferences, while large k introduces artifacts that weaken supervision.

\noindent\textbf{Impact of Timestep Range.}  
We apply EASPO on a subset of denoising steps $[0, \kappa]$ in Tab.~\ref{tab:timestep_range} and find that skipping noisy early steps improves alignment due to limited speech structure, while omitting fine-grained late steps weakens prosodic refinement. A mid-range window (e.g., $[0,750]$) balances diversity and emotional clarity, yielding optimal performance.

\noindent\textbf{Comparison with other diffusion-based RL alignment methods.}
We compare EASPO to prior diffusion-based RL methods (DDPO, D3PO, Diff-DPO) in Tab.~\ref{tab:align_methods}, observing consistent metric gains that highlight EASPO’s strength in aligning fine-grained emotional and prosodic preferences.

\noindent\textbf{Choice of Win/Lose Pairs.}  
We compare selecting the top–bottom EASPM-scored candidates versus randomly sampled pairs from the same latent state in Fig.~\ref{tab:win_lose_choice}. Using highest–lowest scoring samples ensures stronger emotional contrast while maintaining comparable content and noise levels, leading to more stable and informative supervision.
\section{Conclusion}
We present EASPO, a diffusion-based speech synthesis framework that introduces stepwise preference optimization for fine-grained emotional alignment. By leveraging an emotion-aware scoring model to compare candidate samples at each denoising step, EASPO progressively guides generation toward emotionally expressive and prosodically natural speech. This step-conditioned training strategy enables the model to capture subtle affective cues through contrastive supervision. Extensive experimental  demonstrate the effectiveness across both objective and subjective evaluations.

\bibliographystyle{IEEEbib}
\bibliography{strings,main}

\end{document}